\documentclass[letterpaper]{article} 
\usepackage{aaai2026}  
\usepackage{times}  
\usepackage{helvet}  
\usepackage{courier}  
\usepackage[hyphens]{url}  
\usepackage{graphicx} 
\usepackage{natbib}  
\usepackage{caption} 
\usepackage{algorithm}
\usepackage{algorithmic}

\usepackage{newfloat}
\usepackage{listings}
\DeclareCaptionStyle{ruled}{labelfont=normalfont,labelsep=colon,strut=off} 
\floatstyle{ruled}
\newfloat{listing}{tb}{lst}{}
\floatname{listing}{Listing}
\usepackage{booktabs}
\usepackage{amsmath}
\usepackage{subcaption}
\usepackage{multirow} 
\usepackage{threeparttable}
\usepackage{marvosym}
\usepackage{textcomp}
\usepackage{amsfonts}

\title{Length-Adaptive Interest Network for Balancing Long and Short Sequence Modeling in CTR Prediction}

\author{
    Zhicheng Zhang\textsuperscript{\rm 1}\equalcontrib,
    Zhaocheng Du\textsuperscript{\rm 2}\equalcontrib,
    Jieming Zhu\textsuperscript{\rm 2 ${\dagger}$},
    Jiwei Tang\textsuperscript{\rm 1},
    Fengyuan Lu\textsuperscript{\rm 3},
    Wang Jiaheng\textsuperscript{\rm 4},
    \\
    Song-Li Wu\textsuperscript{\rm 1},
    Qianhui Zhu\textsuperscript{\rm 1},
    Jingyu Li\textsuperscript{\rm 5},
    Hai-Tao Zheng\textsuperscript{\rm 1,6}\thanks{Corresponding Authors: Jieming Zhu and Hai-Tao Zheng.},
    Zhenhua Dong\textsuperscript{\rm 2}
}

\affiliations{
    \textsuperscript{\rm 1}Shenzhen International Graduate School, Tsinghua University\\
    \textsuperscript{\rm 2}Huawei Noah's Ark Lab\\
    \textsuperscript{\rm 3}School of Artificial Intelligence and Science, Nanjing University\\
    \textsuperscript{\rm 4}Hong Kong University of Science and Technology\\
    \textsuperscript{\rm 5}School of Cyber Science and Technology, Sun Yat-sen University\\
    \textsuperscript{\rm 6}Peng Cheng Laboratory\\

    zhang-zc24@mails.tsinghua.edu.cn, zhaochengdu@huawei.com, jiemingzhu@ieee.org, zheng.haitao@sz.tsinghua.edu.cn
}

\begin{document}

\maketitle

\begin{abstract}
    User behavior sequences in modern recommendation systems exhibit significant length heterogeneity, ranging from sparse short-term interactions to rich long-term histories. While longer sequences provide more context, we observe that increasing the maximum input sequence length in existing CTR models paradoxically degrades performance for short-sequence users due to attention polarization and length imbalance in training data. 
    To address this, we propose \textbf{LAIN} (Length-Adaptive Interest Network), a plug-and-play
framework that explicitly incorporates sequence length as a conditioning signal to balance long- and short-sequence modeling. LAIN consists of three lightweight components: a \textit{Spectral Length Encoder} that maps length into continuous representations, \textit{Length-Conditioned Prompting} that injects global contextual cues into both long- and short-term behavior branches, and \textit{Length-Modulated Attention} that adaptively adjusts attention sharpness based on sequence length. 
    Extensive experiments on three real-world benchmarks across five strong CTR backbones show that LAIN consistently improves overall performance, achieving up to 1.15\% AUC gain and 2.25\% log loss reduction. Notably, our method significantly improves accuracy for short-sequence users without sacrificing long-sequence effectiveness. 
    Our work offers a general, efficient, and deployable solution to mitigate length-induced bias in sequential recommendation.
\end{abstract}

\section{Introduction}

\begin{figure}[!t]
\centering
  \begin{subfigure}[t]{1\linewidth}
    \centering
    \includegraphics[width=0.95\linewidth]{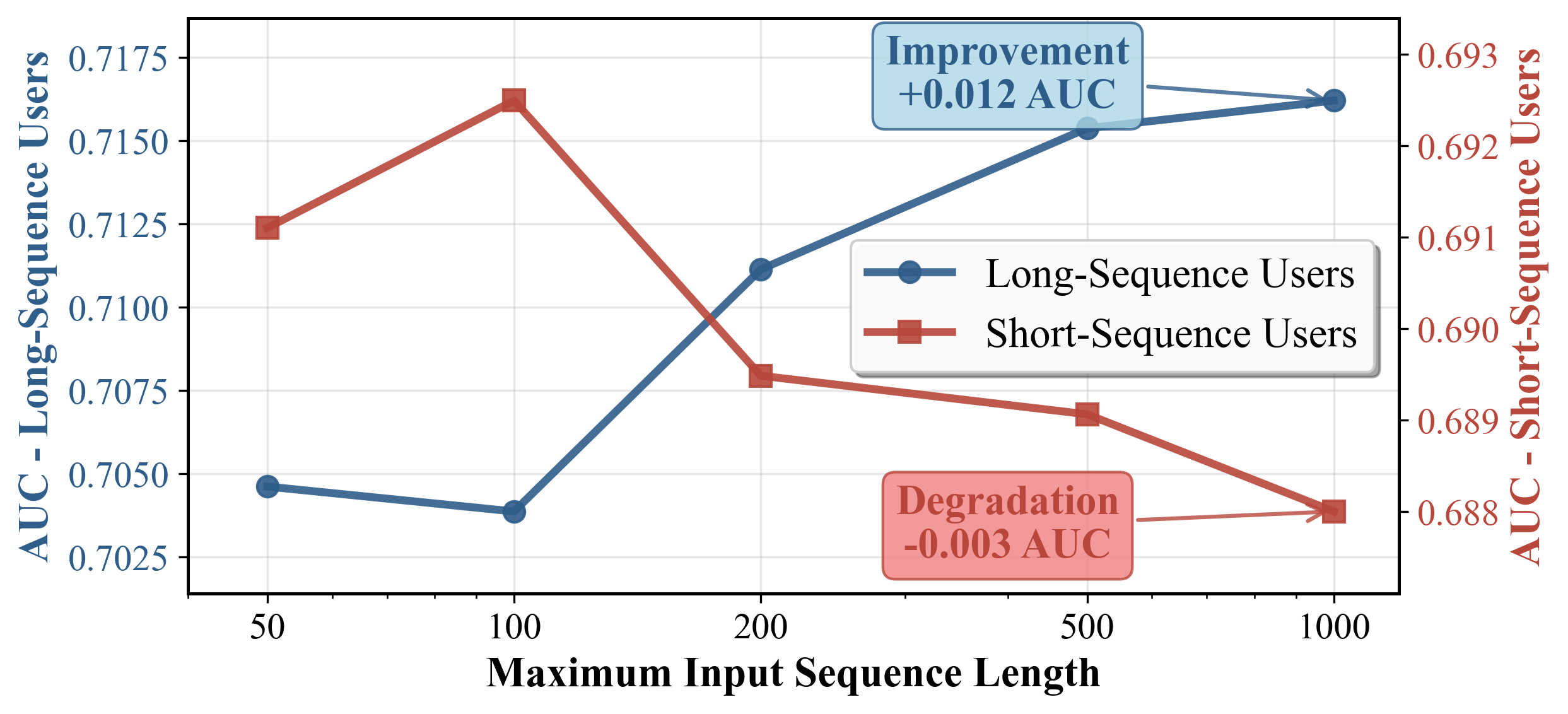}
    \subcaption{Performance Imbalance : Long- vs Short-Sequence Users}
    \label{fig:intro_auc}
  \end{subfigure}
  \hfill
  \begin{subfigure}[t]{0.9\linewidth}
    \centering
    \includegraphics[width=\linewidth]{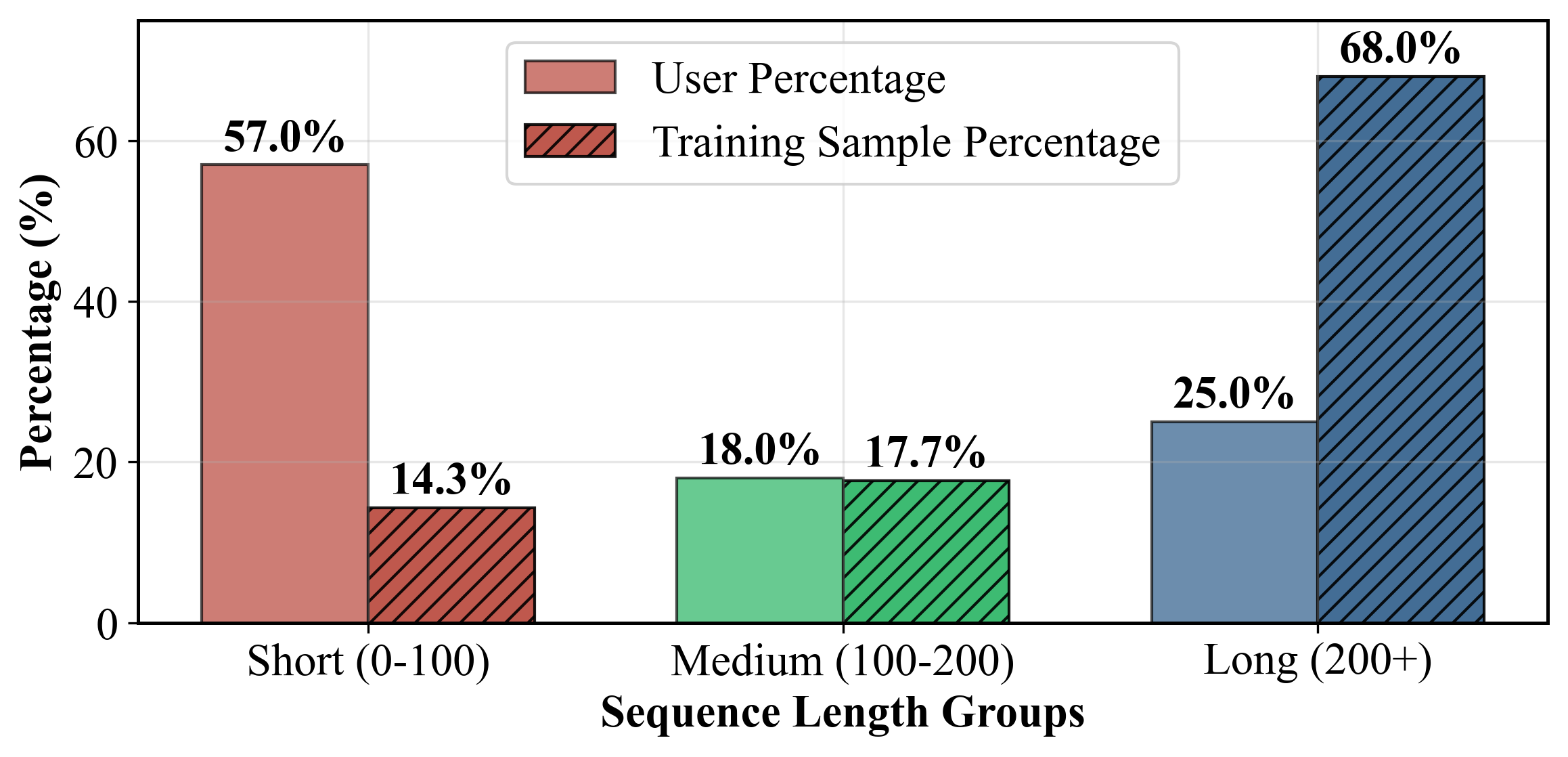}
    \subcaption{Data Imbalance : Users vs Training Samples}
    \label{fig:intro_dist}
  \end{subfigure}
  \caption{
   (a) The effect of varying the maximum input sequence length on average AUC across five CTR models on the \textit{EBNeRD-small} dataset. Longer sequences improve performance for users with extensive behavior histories ($>200$), while consistently degrading it for users with fewer interactions ($<100$). 
   (b) The user behavior length distribution reveals a skewed pattern: short-sequence users form the majority, yet long-sequence users dominate the training sample.
  }
  \label{fig:intro_motivation}
\end{figure}

Click-through rate (CTR) prediction is a core task in personalized recommendation and online advertising systems. Accurately capturing user preferences from historical behavior sequences has shown to significantly improve prediction performance~\cite{zhou2018deep, zhouDeepInterestEvolution2019, 10.1145/3580305.3599331, 10.1145/3711896.3737262}. As user engagement grows and digital content consumption increases, user behavior sequences have become longer and more diverse. Consequently, a growing body of work has focused on long-sequence modeling frameworks~\cite{piSearchbasedUserInterest2020, changTWINTWostageInterest2023a, siTWINV2Scaling2024}, which typically adopt a two-stage architecture: 
a general search unit (GSU) that selects representative behaviors relevant to the target item from the full history, and an exact search unit (ESU) that applies target-aware attention over selected behaviors to derive a refined long-sequence representation. In parallel, a short-sequence branch often extracts the most recent interactions for high-resolution modeling of recent interests. The fused representation is then used for CTR prediction. This framework has been widely deployed in industrial systems, demonstrating its ability to leverage rich contextual information from long behavior sequences.

Despite its success, we identify a counterintuitive phenomenon in deployment: \emph{as the maximum input sequence length increases, the prediction performance for long-sequence users improves, but the performance for short-sequence users degrades notably}. As shown in Figure~\ref{fig:intro_auc}, users with long histories  benefit from longer inputs, whereas users with fewer behaviors suffer from decreased AUC. 
This phenomenon is particularly concerning since short-sequence users often constitute the majority. As illustrated in Figure~\ref{fig:intro_dist}, 57\% of users contribute only 14.3\% of training samples, while just 25\% of users with long sequences dominate over 68\% of training data.

This imbalance poses a practical dilemma. Models are implicitly biased toward optimizing for high-frequency long-sequence users, whose patterns dominate the training data. 
Yet in reality, short-sequence users are more prevalent\cite{10.1145/3394486.3403078}.
Standard CTR models do not explicitly differentiate between users of varying sequence lengths. Instead, they apply a one-size-fits-all architecture across the entire spectrum of sequence lengths, assuming that user behaviors are homogeneously structured~\cite{Mao_Zhu_Su_Cai_Li_Dong_2023,10.1145/3583780.3615134}. 
As a consequence, new or inactive users---a critical cohort for platform growth---may suffer degraded recommendation quality, limiting system-wide engagement and retention.

To address this issue, we conduct an in-depth diagnosis (Section Empirical Analysis) and identify two key factors. The first is \textit{attention polarization}: softmax-based attention tends to over-concentrate on a few salient behaviors in long sequences, which benefits interest extraction but amplifies noise in short sequences with limited context. The second is \textit{length signal deficiency}: existing models treat behavior sequences as homogeneous event sets, failing to leverage sequence length as a prior that reflects user status. Together, these effects impair short-sequence modeling and hinder balanced performance across user groups.

Motivated by these findings, we propose that \emph{sequence length should be explicitly modeled as a conditional signal} to guide interest extraction and attention behavior adaptively. To this end, we introduce the \textbf{Length-Adaptive Interest Network (LAIN)}, a lightweight, plug-and-play framework that injects length-awareness into CTR prediction. LAIN comprises three key components: a \emph{Spectral Length Encoder} that maps raw length values into rich embeddings via continuous Fourier bases; a \emph{Length-Conditioned Prompting} module that generates prompt tokens conditioned on sequence length to modulate user representations; and a \emph{Length-Modulated Attention} mechanism that adaptively adjusts attention sharpness by modulating the temperature of the softmax function and augmenting key-query representations based on length priors.

By design, LAIN is compatible with mainstream CTR models and can be seamlessly integrated into existing architectures with negligible inference overhead. Our approach enables a shared model to dynamically adapt its attention and representation strategies to user sequences of varying lengths. Extensive experiments on multiple large-scale public datasets demonstrate that LAIN consistently improves both overall CTR performance and, crucially, significantly boosts prediction quality for short-sequence users. We believe this study highlights an important yet under-explored dimension of sequence modeling in recommendation systems and provides a generalizable paradigm for length-adaptive user modeling.

Our contributions are summarized as follows:

\begin{itemize}
    \item We identify and analyze the underexplored challenge of length imbalance in CTR modeling and characterize its effects as \textit{attention polarization} and \textit{length signal deficiency}.
    \item We propose LAIN, a novel, lightweight and modular length-aware enhancement framework with three components that explicitly encode, inject, and modulate sequence length in user modeling.
    \item We conduct extensive experiments demonstrating that LAIN substantially improves both long- and short-sequence modeling, and provides a generalizable solution to mitigating length-induced bias in CTR prediction.
\end{itemize}

\section{Related Work}

\paragraph{Long‑sequence CTR models.}  
Modern recommender systems often handle very long user behavior sequences using a two‑stage architecture: a General Search Unit (GSU) retrieves a subset of relevant behaviors, and an Exact Search Unit (ESU) applies target-aware attention for final modeling. Prominent models following this paradigm include SIM~\cite{piSearchbasedUserInterest2020}, ETA~\cite{chen2022efficientlongsequentialuser}, SDIM~\cite{cao2022sdim}, and TWIN~\cite{changTWINTWostageInterest2023a}. The more recent TWIN‑V2 further introduces hierarchical clustering to compress life‑cycle behaviors and allows modeling of ultra‑long sequences in a scalable manner~\cite{siTWINV2Scaling2024}. However, such frameworks focus primarily on expanding sequence length and search efficiency while largely overlooking the performance imbalance between long‑ and short‑sequence users. Recent work DARE~\cite{feng2025longsequence} critiques this by decoupling attention and representation embeddings to mitigate interference, but still does not condition on sequence length during modeling.

\paragraph{Length imbalance and fairness in recommendation.}  
Despite the widespread presence of long‑tail behavior distributions~\cite{10.1145/3583780.3614929}, few prior works explicitly address the modeling imbalance between short- and long‑history users. Traditional CTR models optimize aggregate metrics like AUC or log loss, implicitly favoring data-rich long‑history users, while potentially under‑serving the majority short‑history cohort~\cite{bars_cikm}. Although fairness and cold-start literature address new or infrequent users~\cite{10.1145/3292500.3330859,jiang2024promptcoldstart}, to our knowledge no existing work systematically diagnoses the phenomenon of length imbalance or proposes conditioning the model on sequence length to balance across user groups.

\paragraph{Prompting and conditional encoding in recommendation.}  
Recent research has explored soft prompt and prefix mechanisms to inject auxiliary context into sequential recommendation~\cite{wang2025personalized}. For example, Personalized Prompt for Sequential Recommendation (PPR) generates soft prefix prompts from user profiles to improve cold-start performance~\cite{wu2022personalizedprompt}. Other studies propose custom prompt tuning atop pre-trained recommendation models to enable efficient adaptation~\cite{10.1145/3746252.3761574,10.1145/3589334.3645380}. Unlike these methods, our Length-Conditioned Prompting dynamically generates prompts conditioned on behavior sequence length—not on user profile or item metadata—allowing the model to adapt its attention strategy according to implicit user state.

\section{Empirical Analysis}
\label{sec:ob}

In this section, we conduct an in-depth empirical investigation to reveal why the widely-adopted long-short sequence modeling paradigm may systematically underperform for short-sequence users, despite improving the overall performance. Our diagnosis identifies two key factors: \textit{Attention Polarization} and \textit{Length Signal Deficiency}. These factors indicate that user sequence length is not just a passive statistic, but an active condition that influences model behavior, and must be explicitly modeled.

\paragraph{Attention Polarization}

\begin{figure}[h]
    \centering
    \includegraphics[width=\linewidth]{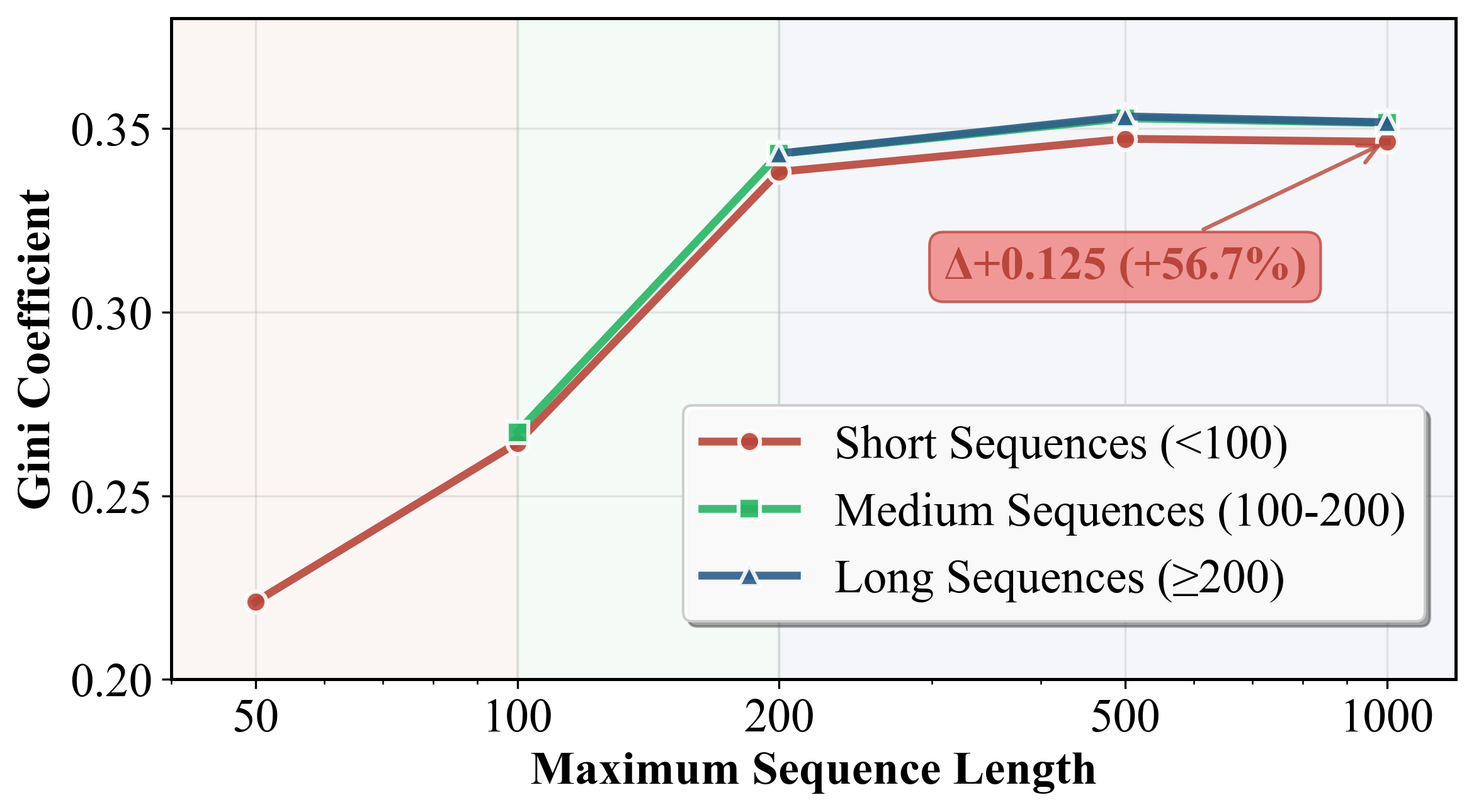}
    \caption{Attention polarization trend in baseline CTR models. Gini coefficient increases dramatically with sequence length, demonstrating progressive attention concentration that particularly affects short-sequence users.}
    \label{fig:attention_gini2}
\end{figure}

Transformer-based CTR models rely on attention mechanisms—particularly target-aware attention—to extract user preferences from historical behavior sequences. However, the use of softmax attention inherently involves a normalization constraint over all tokens in the sequence. As the input sequence grows longer, the softmax distribution becomes more \textit{polarized}, concentrating attention weights on a few "salient" items and suppressing the rest. This effect, visualized via Gini coefficient or entropy measurements, becomes more pronounced with longer input lengths (see Figure~\ref{fig:attention_gini2}).

While this selective focus may benefit long-history users by isolating strong preferences, it proves detrimental for short-history users. When only a few behaviors are available (e.g., $<10$ items), softmax often places overwhelming weight on one or two items, effectively ignoring others. This leads to underutilization of limited signals, resulting in unstable and brittle interest representations. 

Such behavior aligns with findings in other domains, such as language modeling and vision, where attention entropy is negatively correlated with input length~\cite{tang-etal-2025-perception, tang2025gmsaenhancingcontextcompression}. In our CTR setting, this "attention collapse" disproportionately harms short-sequence users and necessitates strategies that regulate attention sharpness as a function of sequence length.

\paragraph{Length Signal Deficiency}
Another critical issue is that most CTR models treat behavior sequences as homogeneous unordered event sets and omit sequence length from the modeling process. This design overlooks the fact that sequence length is a powerful proxy for a user's latent state.

Empirically, we observe strong correlations between sequence length and user behavioral characteristics: (1) \textbf{short-sequence users} tend to be new, cold-start, or low-activity users with unstable interests and lower click-through rates; (2) \textbf{long-sequence users} exhibit higher activity, greater interest diversity, and more stable long-term preferences.

\begin{table}[t]
\centering
\scriptsize
\setlength{\tabcolsep}{3pt}
\renewcommand{\arraystretch}{0.8}
\begin{tabular}{p{2.4cm}|ccc}
\toprule
\textbf{Metric} &
\textbf{Short ($<100$)} &
\textbf{Medium ($100$--$200$)} &
\textbf{Long ($\ge200$)} \\
\midrule
\multicolumn{4}{l}{\textit{User Distribution}} \\
User Count & 10,735 & 3,379 & 4,713 \\
Percentage (\%) & 57.0 & 18.0 & 25.0 \\

\midrule
\multicolumn{4}{l}{\textit{Sequence Characteristics}} \\
Avg Length & 37.1 & 144.2 & 401.4 \\
Length Std & 26.9 & 28.7 & 183.3 \\

\midrule
\multicolumn{4}{l}{\textit{Behavioral Patterns}} \\
Click Rate (\%) & 8.28 & 8.60 & 9.41 \\
Click Rate Std (\%) & 4.19 & 3.04 & 2.52 \\
Behavioral Stability & 0.515 & 0.354 & 0.268 \\

\midrule
\multicolumn{4}{l}{\textit{Content Engagement}} \\
Avg Unique Types & 2.5 & 3.6 & 4.6 \\
Avg Sessions/User & 3.1 & 7.5 & 15.7 \\
Avg Impr./User & 5.5 & 13.4 & 32.0 \\
\bottomrule
\end{tabular}
\caption{Length Signal Deficiency: Empirical Evidence from User Behavioral Analysis. Behavioral Stability measured by Coefficient of Variation of click rates; lower values indicate more stable behavior. Data from EBNeRD-small.}
\label{tab:length_signal_deficiency}
\end{table}

Our comprehensive analysis on the Ebnerd dataset reveals significant disparities across sequence length groups. Short-sequence users constitute 57.0\% of the user base but achieve only 8.28\% click-through rate, while long-sequence users (200+ behaviors) represent 25.0\% of users with 9.41\% click-through rate—a substantial 1.13 percentage point improvement. More critically, short-sequence users exhibit high behavioral instability (coefficient of variation: 0.515) compared to long-sequence users (0.268), indicating that their preferences are indeed more volatile and harder to model consistently.
Furthermore, content engagement patterns differ markedly: short-sequence users interact with an average of 2.5 article types, while long-sequence users engage with 4.6 types, demonstrating the diversity gap mentioned above.

  \begin{figure*}[!t]
      \centering
      \includegraphics[width=2\columnwidth]{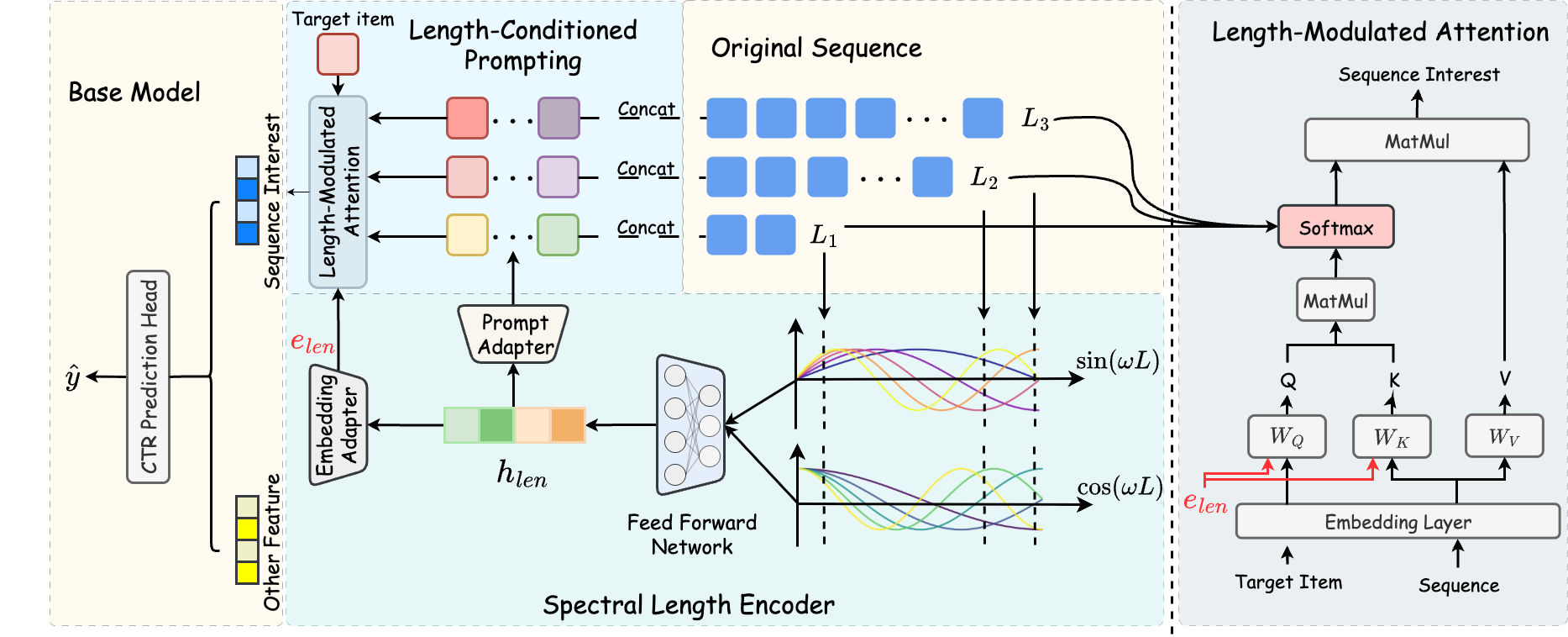}
\caption{Overview of the Length-Adaptive Interest Network (LAIN) for CTR prediction. LAIN conditions on sequence length via a Spectral Length Encoder (SLE), which generates a continuous embedding $\mathbf{h}_{\text{len}}$ from the raw length $L$. This embedding modulates two components: Length-Conditioned Prompting (LCP), which prepends length-aware prompt tokens to the behavior sequence, and Length-Modulated Attention (LMA), which adjusts attention via query/key conditioning and dynamic temperature scaling. The resulting sequence representation is fused with other features for final CTR prediction.}
      \label{fig:model}
  \end{figure*}

Despite this, current models apply the same architecture and attention logic regardless of sequence length, lacking any conditional adaptation. This implicitly assumes that a user with 3 interactions and a user with 300 interactions can be modeled with identical inductive biases. As a result, the model overfits to patterns seen in data-rich long-history users (who dominate the training sample size) and underperforms on the majority of short-history users.

We argue that this is a manifestation of \textbf{length signal deficiency}: failing to condition on sequence length prevents the model from distinguishing between user states at different stages of the behavioral lifecycle. As a result, model capacity is inefficiently allocated, and generalization suffers.

Together, attention polarization and length signal deficiency point to the need for \textit{length-aware modeling}. Models should not only \textbf{observe} sequences but also \textbf{understand how long} they are—and what that implies about the user. 

\section{Method: Length-Adaptive Interest Network}

\paragraph{Preliminaries}

CTR prediction estimates the probability of user-item interactions in recommendation systems, serving as a cornerstone for enhancing user engagement through personalized content delivery and optimizing business outcomes via targeted audience matching. Formulated as a binary classification task, the goal is to learn a predictor $f: \mathbb{R}^d \rightarrow \mathbb{R}$ from training data $\mathcal{D} = \{(\mathbf{x}_i, y_i)\}$, where $\mathbf{x}_i \in \mathbb{R}^d$ concatenates user, item, and contextual features, and $y_i \in \{0, 1\}$ indicates click events. The predicted CTR is:
\begin{equation}
\hat{y}_i = \sigma(\phi(\mathbf{x}_i)),
\end{equation}
with $\sigma(\cdot)$ as the sigmoid function. Training minimizes the binary cross-entropy loss:
\begin{equation}
\mathcal{L} = -\frac{1}{N} \sum_{i=1}^N \left[ y_i \log \hat{y}_i + (1 - y_i) \log(1 - \hat{y}_i) \right],
\end{equation}

The primary challenge for \emph{long-sequence CTR} lies in designing $\phi(\cdot)$ to capture long-range temporal dependencies in user behavior while ensuring computationally efficient training and inference on large-scale sequences.

The architecture for CTR prediction with long sequence modeling follows a structured approach encompassing four core components: \textbf{(1)} feature embedding that maps categorical features to dense vectors; \textbf{(2)} feature interaction modeling to capture cross-field dependencies; \textbf{(3)} short-term interest modeling that extracts recent behavioral patterns; and \textbf{(4)} long-term interest modeling that leverages extended user history through two-stage architectures (GSU + ESU). LAIN enhances this framework by injecting length-awareness into both short-term and long-term interest modeling components.

CTR models typically share a single parameter set $\theta$ to process all sequences regardless of length. However, users with different behavior lengths may exhibit fundamentally different modeling needs. Specifically, short-sequence users benefit from smooth, distributed attention to maximize limited signal utilization, whereas long-sequence users require sparse, concentrated attention to focus on key interests. This induces conflicting gradient signals during training:
\begin{align}
    \nabla_{\theta} \mathcal{L}_{\text{short}} &\propto -\alpha \cdot \text{smoothness gradient},\\
    \nabla_{\theta} \mathcal{L}_{\text{long}} &\propto +\beta \cdot \text{sparsity gradient},\quad \alpha, \beta > 0,
\end{align}
leading to negative inner products $\langle \nabla \mathcal{L}_{\text{short}}, \nabla \mathcal{L}_{\text{long}} \rangle < 0$ and degraded generalization.

LAIN mitigates this by explicitly conditioning the model on sequence length $L$ and decoupling the optimization space via dynamic, length-dependent parameters. Specifically, we decompose model parameters into shared and adaptive components:
\begin{equation}
\theta = \{ \theta_{\text{shared}}, \theta_{\text{prompt}}(L) \},
\end{equation}
where $\theta_{\text{prompt}}(L)$ are generated on-the-fly via a learnable function of $L$ and injected as prompts into the model. The loss becomes:
\begin{equation}
\mathcal{L} = \sum_L \sum_{(u,v_t,y) \in \mathcal{D}_L} \ell(f_{\theta_{\text{shared}}}([\mathbf{S}_u; \mathbf{P}(L)], v_t), y),
\end{equation}
where $\mathbf{P}(L)$ denotes the prompt tokens conditioned on $L$. This decouples gradient flows for users with varying sequence lengths and enables length-specific inductive biases.

\paragraph{Spectral Length Encoder (SLE)}

We first encode the sequence length $L$ into a dense vector via trainable Fourier projections:
\begin{equation}
\mathbf{f}_{\text{fourier}} = [\sin(L \cdot \boldsymbol{\omega}); \cos(L \cdot \boldsymbol{\omega})] \in \mathbb{R}^{2d_f},
\end{equation}
where $\boldsymbol{\omega} \in \mathbb{R}^{d_f}$ are learnable frequency parameters. This representation avoids discretization and captures continuous length semantics~\cite{10.5555/3495724.3496356}. It is projected via an MLP to produce a shared embedding:
\begin{equation}
\mathbf{h}_{\text{len}} = \text{MLP}(\text{LayerNorm}(\text{Linear}(\mathbf{f}_{\text{fourier}}))) \in \mathbb{R}^{d}.
\end{equation}

\paragraph{Length-Conditioned Prompting (LCP)}

We generate $k$ learnable prompt tokens from $\mathbf{h}_{\text{len}}$:
\begin{equation}
\mathbf{P}(L) = \text{reshape}(\text{MLP}_{\text{prompt}}(\mathbf{h}_{\text{len}})) \in \mathbb{R}^{k \times d}.
\end{equation}
These tokens are prepended to both short-term and long-term behavior sequences:
\begin{equation}
\mathbf{S}'_{\text{short}} = [\mathbf{P}; \mathbf{S}_{\text{short}}], \quad \mathbf{S}'_{\text{long}} = [\mathbf{P}; \mathbf{S}_{\text{long}}],
\end{equation}
and then passed to the attention encoder. LCP effectively injects global user state and expands the parameterization space, thus avoiding interference across sequences of varying length.

\paragraph{Length-Modulated Attention (LMA)}

To further improve adaptivity, we modulate the attention computation using $\mathbf{h}_{\text{len}}$.

\paragraph{Query-Key Conditioning.} We concatenate $\mathbf{h}_{\text{len}}$ to each query and key:
\begin{align}
\mathbf{Q}' &= W_q([\mathbf{Q}; \mathbf{e}_{\text{len}}]), \\
\mathbf{K}' &= W_k([\mathbf{K}; \mathbf{e}_{\text{len}}]),
\end{align}
where $\mathbf{e}_{\text{len}} = \text{MLP}_{\text{emb}}(\mathbf{h}_{\text{len}})$ is a length embedding generated by a small MLP.

\paragraph{Softmax Temperature Scaling.} We define a length-aware temperature:
\begin{equation}
\tau = 1 + \text{sigmoid}(-\beta(L - L_0)) \cdot \gamma,
\end{equation}
where $L_0$ is the mean sequence length, and $\gamma$ and $\beta$ are learnable parameters. The attention weights are computed as:
\begin{equation}
\alpha_{ij} = \frac{\exp\left(\frac{\mathbf{Q}'_i \cdot \mathbf{K}'_j}{\sqrt{d} \cdot \tau}\right)}{\sum_j \exp\left(\frac{\mathbf{Q}'_i \cdot \mathbf{K}'_j}{\sqrt{d} \cdot \tau}\right)},
\end{equation}
yielding output:
\begin{equation}
\mathbf{O}_i = \sum_j \alpha_{ij} \cdot \mathbf{V}_j.
\end{equation}
When $L$ is small, $\tau$ increases and smooths the attention distribution, reducing over-polarization. When $L$ is large, $\tau$ shrinks, promoting sharper focus on salient items.

\paragraph{Training and Complexity Analysis}

LAIN is trained end-to-end with minimal overhead. The Spectral Length Encoder (SLE) introduces $\mathcal{O}(d_f d + d^2)$ parameters. Length-Conditioned Prompting (LCP) adds $k \cdot d + \mathcal{O}(k d^2)$ parameters and increases the sequence length from $L$ to $L+k$. Length-Modulated Attention (LMA) contributes $\mathcal{O}(d^2)$ parameters for the conditioned projections and MLP, plus scalar parameters for temperature scaling.

The overall attention complexity is $\mathcal{O}((L+k)^2 d)$. Since $k$ is a small constant ($k=2\sim4$), this simplifies to $\mathcal{O}(L^2 d)$, with only negligible linear terms added. In total, LAIN introduces less than 1.5\% additional parameters and incurs about 2.3\% inference time overhead compared to the base transformer model.

\section{Experiments}

In this section, we conduct extensive experiments to evaluate the effectiveness of our proposed \textit{Length-Conditioned Prompting} framework. 

\paragraph{Datasets.}
We evaluate our method on three real-world sequential CTR datasets: \textbf{KuaiVideo}~\cite{chenTemporalHierarchicalAttention2018a}, \textbf{MicroVideo1.7M}~\cite{chenTemporalHierarchicalAttention2018a}, and \textbf{EBNeRD-small}~\cite{10.1145/3687151.3687152}. These datasets span diverse domains—short videos and news—and exhibit long-tailed user sequence distributions, making them suitable for evaluating length-aware recommendation methods. The datasets statistics are shown in Table ~\ref{tab:dataset_stat}.

\begin{table}[ht]
\centering
\scalebox{0.8}{ 
\begin{tabular}{lrrrr}
\toprule
Dataset & Users & Items & Interactions & Avg-Len \\
\midrule
MicroVideo1.7M & 10,951 & 1,704,880 & 4,287,008 & 148 \\
KuaiVideo & 10,000 & 3,239,534 & 4,664,549 & 278 \\
EBNeRD-small & 18,828 & 20,739 & 5,514,689 & 147 \\
\bottomrule
\end{tabular}
}
\caption{Datasets statistics.}
\label{tab:dataset_stat}
\end{table}

\begin{table*}[ht]
\centering
\begin{tabular}{lcccccccccc}
\toprule
\textbf{Model} & \multicolumn{3}{c}{\textbf{EBNeRD-small}} & \multicolumn{3}{c}{\textbf{KuaiVideo}} & \multicolumn{3}{c}{\textbf{MicroVideo1.7M}} \\
\cmidrule(lr){2-4} \cmidrule(lr){5-7} \cmidrule(lr){8-10}
 & \textbf{GAUC↑} & \textbf{AUC↑} & \textbf{logloss↓} & \textbf{GAUC↑} & \textbf{AUC↑} & \textbf{logloss↓} & \textbf{GAUC↑} & \textbf{AUC↑} & \textbf{logloss↓} \\
\midrule
DIN      & 0.7053 & 0.7080 & 0.2695 & 0.6716 & 0.6979 & 0.4498 & 0.7023 & 0.7093 & 0.4196 \\
\quad +LAIN    & 0.7096 & 0.7156 & 0.2678 & 0.6742 & 0.7043 & 0.4457 & 0.7019 & 0.7101 & 0.4179 \\
Rel. Gain & \textbf{0.61\%} & \textbf{1.07\%} & \textbf{-0.65\%} & \textbf{0.40\%} & \textbf{0.93\%} & \textbf{-0.92\%} & -0.05\% & \textbf{0.12\%} & \textbf{-0.40\%} \\
\midrule
DIEN     & 0.7090 & 0.7099 & 0.2737 & 0.6678 & 0.6977 & 0.4509 & 0.7143 & 0.7185 & 0.4171 \\
\quad +LAIN    & 0.7078 & 0.7118 & 0.2676 & 0.6688 & 0.6991 & 0.4476 & 0.7146 & 0.7219 & 0.4161 \\
Rel. Gain & -0.17\% & \textbf{0.27\%} & \textbf{-2.25\%} & \textbf{0.14\%} & \textbf{0.20\%} & \textbf{-0.71\%} & \textbf{0.04\%} & \textbf{0.47\%} & \textbf{-0.23\%} \\
\midrule
SIM      & 0.6960 & 0.6992 & 0.2720 & 0.6672 & 0.6875 & 0.4574 & 0.7017 & 0.7095 & 0.4168 \\
\quad +LAIN    & 0.6990 & 0.7037 & 0.2706 & 0.6678 & 0.6896 & 0.4566 & 0.7070 & 0.7131 & 0.4151 \\
Rel. Gain & \textbf{0.43\%} & \textbf{0.65\%} & \textbf{-0.50\%} & \textbf{0.08\%} & \textbf{0.31\%} & \textbf{-0.16\%} & \textbf{0.75\%} & \textbf{0.50\%} & \textbf{-0.41\%} \\
\midrule
SDIM     & 0.7039 & 0.7099 & 0.2719 & 0.6729 & 0.6924 & 0.4536 & 0.6984 & 0.7161 & 0.4129 \\
\quad +LAIN    & 0.7099 & 0.7123 & 0.2704 & 0.6732 & 0.6949 & 0.4525 & 0.6993 & 0.7163 & 0.4121 \\
Rel. Gain & \textbf{0.85\%} & \textbf{0.34\%} & \textbf{-0.56\%} & \textbf{0.05\%} & \textbf{0.35\%} & \textbf{-0.23\%} & \textbf{0.13\%} & \textbf{0.03\%} & \textbf{-0.18\%} \\
\midrule
TWIN     & 0.6930 & 0.6993 & 0.2718 & 0.6729 & 0.6918 & 0.4530 & 0.7060 & 0.7158 & 0.4164 \\
\quad +LAIN    & 0.7012 & 0.7074 & 0.2698 & 0.6748 & 0.6976 & 0.4508 & 0.7093 & 0.7233 & 0.4097 \\
Rel. Gain & \textbf{1.19\%} & \textbf{1.15\%} & \textbf{-0.73\%} & \textbf{0.29\%} & \textbf{0.84\%} & \textbf{-0.48\%} & \textbf{0.47\%} & \textbf{1.05\%} & \textbf{-1.63\%} \\
\bottomrule
\end{tabular}
\caption{Overall performance comparison on three real-world datasets. 
``+LAIN'' denotes the model enhanced with our LAIN method,
    and ``Rel. Gain'' reports the relative improvement (\%) over the base model. $\uparrow$ indicates higher is better; $\downarrow$ indicates lower is better.}
\label{tab:main_results}
\end{table*}

\subsection{Experimental Setup}

\paragraph{Baselines.}
We compare LAIN-enhanced models against five representative long-sequence CTR baselines that cover the spectrum of attention-based user modeling approaches: \textbf{(1)} DIN~\cite{zhouDeepInterestNetwork2018a} uses target-aware attention for interest extraction; \textbf{(2)} DIEN~\cite{zhouDeepInterestEvolution2019} models interest evolution with GRU and attention; \textbf{(3)} SIM~\cite{piSearchbasedUserInterest2020} introduces two-stage architecture with general and exact search units; \textbf{(4)} SDIM~\cite{cao2022sdim} enhances SIM with multi-head attention; and \textbf{(5)} TWIN~\cite{changTWINTWostageInterest2023a} employs twin towers for long-sequence modeling. All baselines are implemented in the FuxiCTR\footnote{https://github.com/reczoo/FuxiCTR} framework with identical feature engineering and hyperparameter settings for fair comparison~\cite{bars_sigir}.

\paragraph{Implementation.}

Following standard CTR prediction protocols, we configure all models with maximum sequence length of 1000 to handle long user histories. Feature embeddings use dimension 64, and models are optimized using Adam optimizer with learning rate 0.001 and early stopping based on validation AUC to prevent overfitting. The CTR prediction head employs a simple MLP with ReLU activation and dropout (0.2) for regularization. For the Spectral Length Encoder, we set Fourier dimension $d_f=32$ and use a 2-layer MLP with hidden dimension 512 for length embedding projection. The Length-Conditioned Prompting module generates $k=4$ prompt tokens with embedding dimension matching the backbone model (64). For Length-Modulated Attention, we initialize temperature scaling parameters $\gamma=0.5$ and $\beta=0.01$, with $L_0$ set as the dataset mean sequence length. The query-key conditioning uses a single linear layer to project concatenated features.

\paragraph{Metrics.}
We employ traditional CTR metrics to comprehensively evaluate LAIN's effectiveness. For overall CTR performance, we report: \textbf{(1)} AUC (Area Under ROC Curve) measuring ranking quality; \textbf{(2)} GAUC (Group AUC) computing per-user AUC then averaging to mitigate high-traffic user dominance; and \textbf{(3)} LogLoss measuring calibration quality for probabilistic predictions.

\subsection{Overall Performance}

Table~\ref{tab:main_results} shows that integrating LAIN into each backbone model leads to consistent improvements on all datasets. Notably, GAUC increases by up to +1.2\% and logloss drops by up to -1.6\%, demonstrating that LAIN enhances sequence modeling capacity without sacrificing robustness. Variance arises from differing average sequence-length distributions and attention architectures; gains are larger for two-stage attention backbones (SIM/SDIM/TWIN) that expose this imbalance more clearly.

\subsection{Length-Specific Evaluation}

We further analyze TWIN’s performance across short, medium, and long sequences. Table~\ref{tab:length_results} shows that LAIN achieves consistent gains across all buckets, especially for short sequences ($<$100), where AUC improves by +1.08\% and logloss reduces by -2.17\%. This supports our hypothesis that prompt-based length conditioning better adapts to both cold-start and long-history users by dynamically adjusting attention focus.

\begin{table*}[h]
    \centering
    \begin{tabular}{l|c|c|c|c|c|c|c|c|c}
    \toprule
    \multirow{2}{*}{Length} & \multicolumn{3}{c|}{TWIN} & \multicolumn{3}{c|}{+LAIN} & \multicolumn{3}{c}{Improvement} \\
    \cmidrule{2-10}
     & GAUC & AUC & Logloss & GAUC & AUC & Logloss & $\Delta$GAUC & $\Delta$AUC & $\Delta$Logloss \\
    \midrule
    0-100 & 0.6951 & 0.7108 & 0.4120 & 0.6986 & 0.7184 & 0.4031 & +0.51\% & +1.08\% & -2.17\% \\
    100-200 & 0.7068 & 0.7212 & 0.4037 & 0.7100 & 0.7248 & 0.3986 & +0.45\% & +0.50\% & -1.26\% \\
    200+ & 0.7133 & 0.7125 & 0.4395 & 0.7166 & 0.7238 & 0.4319 & +0.46\% & +1.58\% & -1.74\% \\
    \midrule
    Overall & 0.7060 & 0.7158 & 0.4164 & 0.7093 & 0.7233 & 0.4097 & +0.47\% & +1.05\% & -1.63\% \\
    \bottomrule
    \end{tabular}
    \caption{Overall performance of TWIN(MicroVideo1.7M) with and without LAIN across different sequence lengths.}
    \label{tab:length_results}
\end{table*}

\subsection{Attention Polarization Mitigation}

We quantify attention polarization by computing the Gini coefficient of attention scores across sequences of varying lengths. As demonstrated in our empirical analysis (Section 2), baseline models exhibit significant attention polarization that increases with maximum sequence length, particularly affecting short-sequence users.

Table~\ref{tab:attention_polarization_mitigation} provides a comprehensive quantitative analysis of how LAIN mitigates this attention polarization problem. We compare the Gini coefficients across different sequence length groups for baseline models configured with varying maximum sequence lengths (200, 500, 1000) against our proposed LAIN model. These results verify that LAIN effectively mitigates over-concentration in softmax attention—an issue exacerbated by sequence length—while maintaining the model's capacity to focus on relevant behavioral signals.

\begin{table}[t]
\centering
\scriptsize
\setlength{\tabcolsep}{3pt}
\renewcommand{\arraystretch}{0.9}
\begin{tabular}{p{2.2cm}|ccc}
\toprule
\textbf{Model Configuration} &
\textbf{Short ($<100$)} &
\textbf{Medium ($100$--$200$)} &
\textbf{Long ($\ge200$)} \\
\midrule
\multicolumn{4}{l}{\textit{Baseline Models (Different Max Sequence Lengths)}} \\
Baseline (L=200) & 0.338 & 0.343 & 0.343 \\
Baseline (L=500) & 0.347 & 0.353 & 0.353 \\
Baseline (L=1000) & 0.346 & 0.351 & 0.352 \\
\midrule
\multicolumn{4}{l}{\textit{Proposed LAIN Model}} \\
LAIN & 0.318 & 0.322 & 0.321 \\
\midrule
\multicolumn{4}{l}{\textit{Improvement Analysis}} \\
Variance Reduction & \multicolumn{3}{c}{\textbf{50.6\%} (vs.\ worst baseline)} \\
Range Reduction & \multicolumn{3}{c}{\textbf{33.3\%} (vs.\ worst baseline)} \\
\bottomrule
\end{tabular}
\caption{Attention Polarization Mitigation Analysis: LAIN vs.\ Baseline Models. Lower Gini scores indicate less attention polarization. LAIN exhibits a more uniform attention distribution across sequence lengths.}
\label{tab:attention_polarization_mitigation}
\end{table}

\subsection{Ablation Study}
We conduct an ablation study on TWIN(MicroVideo1.7M) to examine the effect of each LAIN component. As shown in Table~\ref{tab:ablation}, removing any module degrades performance, confirming that all parts contribute to overall effectiveness. Excluding the Length-Modulated Attention (LMA), which couples temperature scaling with query-key conditioning, leads to the largest drop, indicating its importance in stabilizing attention. While temperature scaling alone slightly hurts performance, it shows strong synergy with length-conditioned prompting and query-key conditioning. Removing the short-term branch—where LAIN is applied to short behavior sequences—also reduces accuracy, highlighting the benefit of incorporating length awareness even in short contexts.

\begin{table}[t]
    \centering
    \setlength{\tabcolsep}{3pt} 
    \renewcommand{\arraystretch}{0.95} 
    \begin{tabular}{l|c|c|c}
    \toprule
    Variant & GAUC & AUC & Logloss \\
    \midrule
    LAIN (Full) & \textbf{0.7093} & \textbf{0.7233} & \textbf{0.4097} \\
    w/o LCP & 0.7083 & 0.7228 & 0.4107 \\
    w/o Query-Key Conditioning & 0.7071 & 0.7195 & 0.4148 \\
    w/o Temperature Scaling & 0.7082 & 0.7212 & 0.4111 \\
    w/o LMA & 0.7076 & 0.7189 & 0.4157 \\
    w/o Short-term Branch & 0.7077 & 0.7226 & 0.4137 \\
    \bottomrule
    \end{tabular}
    \caption{Ablation study on TWIN (MicroVideo1.7M).}
    \label{tab:ablation}
\end{table}

\subsection{Parameter Sensitivity Analysis}

\begin{figure}[htbp]
    \centering
    \includegraphics[width=0.4\textwidth]{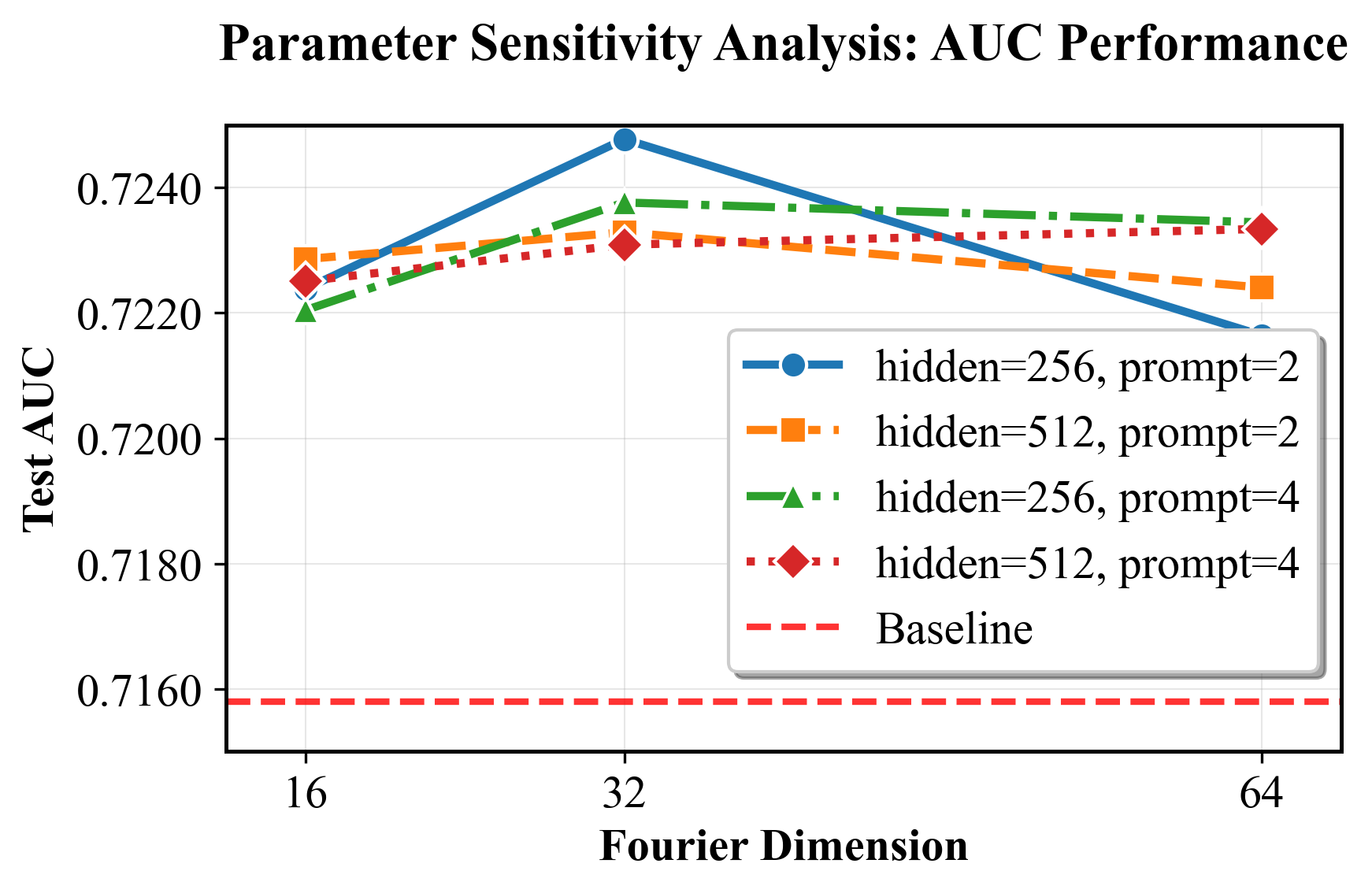}
    \caption{Parameter sensitivity analysis for AUC performance. All hyperparameter configurations consistently outperform the baseline (red dashed line), demonstrating the robustness of our approach across different parameter settings.}
    \label{fig:parameter_interactions}
\end{figure}
We conducted a sensitivity study by jointly varying three major hyperparameters—Fourier feature dimension, hidden dimension, and the number of prompt tokens—across 12 configurations.

As shown in Figure~\ref{fig:parameter_interactions}, all 12 hyperparameter configurations achieve AUC scores ranging from 0.7216 to 0.7248, significantly outperforming the baseline (0.7158). This demonstrates the robustness of our method across different parameter choices. Our selected configuration (64, 512, 4) achieves competitive performance with 1.05\% improvement over baseline, validating our hyperparameter selection.

\section{Conclusion}

In this paper, we present \textbf{LAIN}, a length-adaptive framework for CTR prediction that explicitly incorporates sequence length as a conditioning signal. By introducing three complementary components—Spectral Length Encoder, Length-Conditioned Prompting, and Length-Modulated Attention—LAIN enables adaptive modeling across users with diverse behavior histories. Our method addresses critical challenges such as attention polarization and gradient conflicts caused by length imbalance. Extensive experiments on multiple real-world datasets and strong baselines demonstrate that LAIN consistently improves overall performance, while significantly enhancing CTR prediction for short-sequence users without sacrificing long-sequence accuracy. We believe that LAIN provides a generalizable and practical solution for balanced interest modeling in large-scale recommendation systems.

\section{Acknowledgments}
This research is supported by National Natural Science Foundation of China (Grant No.62276154), Research Center for Computer Network (Shenzhen) Ministry of Education, the Natural Science Foundation of Guangdong Province (Grant No. 2024TQ08X729 and 2023A1515012914), Basic Research Fund of Shenzhen City (Grant No. JCYJ20210324120012033, JCYJ20240813112009013 and GJHZ20240218113603006), the Major Key Projectof PCL for Experiments and Applications (PCL2023A09). 
We also acknowledge partial support from MindSpore, a new deep learning computing framework (\url{https://www.mindspore.cn}).

\bibliography{aaai2026}

@inproceedings{zhou2018deep,
  title={Deep interest network for click-through rate prediction},
  author={Zhou, Guorui and Zhu, Xiaoqiang and Song, Chenru and Fan, Ying and Zhu, Han and Ma, Xiufeng and Yan, Yanghui and Jin, Junqi and Li, Han and Gai, Keping},
  booktitle={Proceedings of the 24th ACM SIGKDD International Conference on Knowledge Discovery \& Data Mining},
  pages={1059--1068},
  year={2018}
}

@article{zhouDeepInterestEvolution2019,
  title = {Deep {{Interest Evolution Network}} for {{Click-Through Rate Prediction}}},
  author = {Zhou, Guorui and Mou, Na and Fan, Ying and Pi, Qi and Bian, Weijie and Zhou, Chang and Zhu, Xiaoqiang and Gai, Kun},
  date = {2019-07-17},
year = {2019},
  journaltitle = {Proceedings of the AAAI Conference on Artificial Intelligence},
  shortjournal = {AAAI},
  volume = {33},
  number = {01},
  pages = {5941--5948},
  issn = {2374-3468, 2159-5399},
  doi = {10.1609/aaai.v33i01.33015941},
  url = {https://ojs.aaai.org/index.php/AAAI/article/view/4545},
  langid = {english}
}

@inproceedings{
feng2025longsequence,
title={Long-Sequence Recommendation Models Need Decoupled Embeddings},
author={Ningya Feng and Junwei Pan and Jialong Wu and Baixu Chen and Ximei Wang and QianLi and Xian Hu and Jie Jiang and Mingsheng Long},
booktitle={The Thirteenth International Conference on Learning Representations},
year={2025},
url={https://openreview.net/forum?id=jkpGIxSsUD}
}

@inproceedings{piSearchbasedUserInterest2020,
  title = {Search-Based {{User Interest Modeling}} with {{Lifelong Sequential Behavior Data}} for {{Click-Through Rate Prediction}}},
  booktitle = {Proceedings of the 29th {{ACM International Conference}} on {{Information}} \& {{Knowledge Management}}},
  author = {Pi, Qi and Zhou, Guorui and Zhang, Yujing and Wang, Zhe and Ren, Lejian and Fan, Ying and Zhu, Xiaoqiang and Gai, Kun},
  date = {2020-10-19},
  year = {2020},
  pages = {2685--2692},
  publisher = {ACM},
  location = {Virtual Event Ireland},
  doi = {10.1145/3340531.3412744},
  url = {https://dl.acm.org/doi/10.1145/3340531.3412744},
  eventtitle = {{{CIKM}} '20: {{The}} 29th {{ACM International Conference}} on {{Information}} and {{Knowledge Management}}},
  isbn = {978-1-4503-6859-9},
  langid = {english}
}

@inproceedings{changTWINTWostageInterest2023a,
  title = {{{TWIN}}: {{TWo-stage Interest Network}} for {{Lifelong User Behavior Modeling}} in {{CTR Prediction}} at {{Kuaishou}}},
  shorttitle = {{{TWIN}}},
  booktitle = {Proceedings of the 29th {{ACM SIGKDD Conference}} on {{Knowledge Discovery}} and {{Data Mining}}},
  author = {Chang, Jianxin and Zhang, Chenbin and Fu, Zhiyi and Zang, Xiaoxue and Guan, Lin and Lu, Jing and Hui, Yiqun and Leng, Dewei and Niu, Yanan and Song, Yang and Gai, Kun},
  date = {2023-08-06},
  year = {2023},
  pages = {3785--3794},
  publisher = {ACM},
  location = {Long Beach CA USA},
  doi = {10.1145/3580305.3599922},
  url = {https://dl.acm.org/doi/10.1145/3580305.3599922},
  eventtitle = {{{KDD}} '23: {{The}} 29th {{ACM SIGKDD Conference}} on {{Knowledge Discovery}} and {{Data Mining}}},
  isbn = {979-8-4007-0103-0},
  langid = {english}
}

@inproceedings{siTWINV2Scaling2024,
  title = {{{TWIN V2}}: {{Scaling Ultra-Long User Behavior Sequence Modeling}} for {{Enhanced CTR Prediction}} at {{Kuaishou}}},
  shorttitle = {{{TWIN V2}}},
  booktitle = {Proceedings of the 33rd {{ACM International Conference}} on {{Information}} and {{Knowledge Management}}},
  author = {Si, Zihua and Guan, Lin and Sun, Zhongxiang and Zang, Xiaoxue and Lu, Jing and Hui, Yiqun and Cao, Xingchao and Yang, Zeyu and Zheng, Yichen and Leng, Dewei and Zheng, Kai and Zhang, Chenbin and Niu, Yanan and Song, Yang and Gai, Kun},
  date = {2024-10-21},
  year = {2024},
  pages = {4890--4897},
  publisher = {ACM},
  location = {Boise ID USA},
  doi = {10.1145/3627673.3680030},
  url = {https://dl.acm.org/doi/10.1145/3627673.3680030},
  eventtitle = {{{CIKM}} '24: {{The}} 33rd {{ACM International Conference}} on {{Information}} and {{Knowledge Management}}},
  isbn = {979-8-4007-0436-9},
  langid = {english}
}

@misc{chen2022efficientlongsequentialuser,
      title={Efficient Long Sequential User Data Modeling for Click-Through Rate Prediction}, 
      author={Qiwei Chen and Yue Xu and Changhua Pei and Shanshan Lv and Tao Zhuang and Junfeng Ge},
      year={2022},
      eprint={2209.12212},
      archivePrefix={arXiv},
      primaryClass={cs.IR},
      url={https://arxiv.org/abs/2209.12212}, 
}

@inproceedings{cao2022sdim,
author = {Cao, Yue and Zhou, Xiaojiang and Feng, Jiaqi and Huang, Peihao and Xiao, Yao and Chen, Dayao and Chen, Sheng},
title = {Sampling Is All You Need on Modeling Long-Term User Behaviors for CTR Prediction},
year = {2022},
isbn = {9781450392365},
publisher = {Association for Computing Machinery},
address = {New York, NY, USA},
url = {https://doi.org/10.1145/3511808.3557082},
doi = {10.1145/3511808.3557082},
booktitle = {Proceedings of the 31st ACM International Conference on Information \& Knowledge Management},
pages = {2974–2983},
numpages = {10},
location = {Atlanta, GA, USA},
series = {CIKM '22}
}

@inproceedings{jiang2024promptcoldstart,
author = {Jiang, Yuezihan and Chen, Gaode and Zhang, Wenhan and Wang, Jingchi and Jiang, Yinjie and Zhang, Qi and Lin, Jingjian and Jiang, Peng and Bian, Kaigui},
title = {Prompt Tuning for Item Cold-start Recommendation},
year = {2024},
isbn = {9798400705052},
publisher = {Association for Computing Machinery},
address = {New York, NY, USA},
url = {https://doi.org/10.1145/3640457.3688126},
doi = {10.1145/3640457.3688126},
booktitle = {Proceedings of the 18th ACM Conference on Recommender Systems},
pages = {411–421},
numpages = {11},
location = {Bari, Italy},
series = {RecSys '24}
}

@article{wu2022personalizedprompt,
author = {Wu, Yiqing and Xie, Ruobing and Zhu, Yongchun and Zhuang, Fuzhen and Zhang, Xu and Lin, Leyu and He, Qing},
title = {Personalized Prompt for Sequential Recommendation},
year = {2024},
issue_date = {July 2024},
publisher = {IEEE Educational Activities Department},
address = {USA},
volume = {36},
number = {7},
issn = {1041-4347},
url = {https://doi.org/10.1109/TKDE.2024.3357498},
doi = {10.1109/TKDE.2024.3357498},
journal = {IEEE Trans. on Knowl. and Data Eng.},
month = jul,
pages = {3376–3389},
numpages = {14}
}

@inproceedings{10.1145/3589334.3645380,
author = {Han, Yongqiang and Wang, Hao and Wang, Kefan and Wu, Likang and Li, Zhi and Guo, Wei and Liu, Yong and Lian, Defu and Chen, Enhong},
title = {Efficient Noise-Decoupling for Multi-Behavior Sequential Recommendation},
year = {2024},
isbn = {9798400701719},
publisher = {Association for Computing Machinery},
address = {New York, NY, USA},
url = {https://doi.org/10.1145/3589334.3645380},
doi = {10.1145/3589334.3645380},
booktitle = {Proceedings of the ACM Web Conference 2024},
pages = {3297–3306},
numpages = {10},
location = {Singapore, Singapore},
series = {WWW '24}
}

@inproceedings{tang-etal-2025-perception,
    title = "Perception Compressor: A Training-Free Prompt Compression Framework in Long Context Scenarios",
    author = "Tang, Jiwei  and
      Xu, Jin  and
      Lu, Tingwei  and
      Zhang, Zhicheng  and
      YimingZhao, YimingZhao  and
      LinHai, LinHai  and
      Zheng, Hai-Tao",
    editor = "Chiruzzo, Luis  and
      Ritter, Alan  and
      Wang, Lu",
    booktitle = "Findings of the Association for Computational Linguistics: NAACL 2025",
    month = apr,
    year = "2025",
    address = "Albuquerque, New Mexico",
    publisher = "Association for Computational Linguistics",
    url = "https://aclanthology.org/2025.findings-naacl.229/",
    doi = "10.18653/v1/2025.findings-naacl.229",
    pages = "4093--4108",
    ISBN = "979-8-89176-195-7"
}

@misc{tang2025gmsaenhancingcontextcompression,
      title={GMSA: Enhancing Context Compression via Group Merging and Layer Semantic Alignment}, 
      author={Jiwei Tang and Zhicheng Zhang and Shunlong Wu and Jingheng Ye and Lichen Bai and Zitai Wang and Tingwei Lu and Jiaqi Chen and Lin Hai and Hai-Tao Zheng and Hong-Gee Kim},
      year={2025},
      eprint={2505.12215},
      archivePrefix={arXiv},
      primaryClass={cs.CL},
      url={https://arxiv.org/abs/2505.12215}, 
}

@inproceedings{chenTemporalHierarchicalAttention2018a,
  title = {Temporal {{Hierarchical Attention}} at {{Category-}} and {{Item-Level}} for {{Micro-Video Click-Through Prediction}}},
  booktitle = {Proceedings of the 26th {{ACM}} International Conference on {{Multimedia}}},
  author = {Chen, Xusong and Liu, Dong and Zha, Zheng-Jun and Zhou, Wengang and Xiong, Zhiwei and Li, Yan},
  date = {2018-10-15},
  year = {2018},
  pages = {1146--1153},
  publisher = {ACM},
  location = {Seoul Republic of Korea},
  doi = {10.1145/3240508.3240617},
  url = {https://dl.acm.org/doi/10.1145/3240508.3240617},
  eventtitle = {{{MM}} '18: {{ACM Multimedia Conference}}},
  isbn = {978-1-4503-5665-7},
  langid = {english}
}

@inproceedings{10.1145/3687151.3687152,
author = {Kruse, Johannes and Lindskow, Kasper and Kalloori, Saikishore and Polignano, Marco and Pomo, Claudio and Srivastava, Abhishek and Uppal, Anshuk and Andersen, Michael Riis and Frellsen, Jes},
title = {EB-NeRD a large-scale dataset for news recommendation},
year = {2024},
isbn = {9798400711275},
publisher = {Association for Computing Machinery},
address = {New York, NY, USA},
url = {https://doi.org/10.1145/3687151.3687152},
doi = {10.1145/3687151.3687152},
booktitle = {Proceedings of the Recommender Systems Challenge 2024},
pages = {1–11},
numpages = {11},
location = {Bari, Italy},
series = {RecSysChallenge '24}
}

@inproceedings{zhouDeepInterestNetwork2018a,
  title = {Deep {{Interest Network}} for {{Click-Through Rate Prediction}}},
  booktitle = {Proceedings of the 24th {{ACM SIGKDD International Conference}} on {{Knowledge Discovery}} \& {{Data Mining}}},
  author = {Zhou, Guorui and Zhu, Xiaoqiang and Song, Chenru and Fan, Ying and Zhu, Han and Ma, Xiao and Yan, Yanghui and Jin, Junqi and Li, Han and Gai, Kun},
  date = {2018-07-19},
year = {2018},
  pages = {1059--1068},
  publisher = {ACM},
  location = {London United Kingdom},
  doi = {10.1145/3219819.3219823},
  url = {https://dl.acm.org/doi/10.1145/3219819.3219823},
  eventtitle = {{{KDD}} '18: {{The}} 24th {{ACM SIGKDD International Conference}} on {{Knowledge Discovery}} and {{Data Mining}}},
  isbn = {978-1-4503-5552-0},
  langid = {english}
}

@inproceedings{10.1145/3394486.3403078,
author = {Yin, Jianwen and Liu, Chenghao and Wang, Weiqing and Sun, Jianling and Hoi, Steven C.H.},
title = {Learning Transferrable Parameters for Long-tailed Sequential User Behavior Modeling},
year = {2020},
isbn = {9781450379984},
publisher = {Association for Computing Machinery},
address = {New York, NY, USA},
url = {https://doi.org/10.1145/3394486.3403078},
doi = {10.1145/3394486.3403078},
booktitle = {Proceedings of the 26th ACM SIGKDD International Conference on Knowledge Discovery \& Data Mining},
pages = {359–367},
numpages = {9},
location = {Virtual Event, CA, USA},
series = {KDD '20}
}

@inproceedings{10.5555/3495724.3496356,
author = {Tancik, Matthew and Srinivasan, Pratul P. and Mildenhall, Ben and Fridovich-Keil, Sara and Raghavan, Nithin and Singhal, Utkarsh and Ramamoorthi, Ravi and Barron, Jonathan T. and Ng, Ren},
title = {Fourier features let networks learn high frequency functions in low dimensional domains},
year = {2020},
isbn = {9781713829546},
publisher = {Curran Associates Inc.},
address = {Red Hook, NY, USA},
articleno = {632},
numpages = {11},
location = {Vancouver, BC, Canada},
series = {NIPS '20}
}

@inproceedings{bars_sigir,
author = {Zhu, Jieming and Dai, Quanyu and Su, Liangcai and Ma, Rong and Liu, Jinyang and Cai, Guohao and Xiao, Xi and Zhang, Rui},
title = {BARS: Towards Open Benchmarking for Recommender Systems},
year = {2022},
isbn = {9781450387323},
publisher = {Association for Computing Machinery},
address = {New York, NY, USA},
url = {https://doi.org/10.1145/3477495.3531723},
doi = {10.1145/3477495.3531723},
booktitle = {Proceedings of the 45th International ACM SIGIR Conference on Research and Development in Information Retrieval},
pages = {2912–2923},
numpages = {12},
location = {Madrid, Spain},
series = {SIGIR '22}
}

@inproceedings{bars_cikm,
  author       = {Jieming Zhu and
                  Jinyang Liu and
                  Shuai Yang and
                  Qi Zhang and
                  Xiuqiang He},
  editor       = {Gianluca Demartini and
                  Guido Zuccon and
                  J. Shane Culpepper and
                  Zi Huang and
                  Hanghang Tong},
  title        = {Open Benchmarking for Click-Through Rate Prediction},
  booktitle    = {{CIKM} '21: The 30th {ACM} International Conference on Information
                  and Knowledge Management, Virtual Event, Queensland, Australia, November
                  1 - 5, 2021},
  pages        = {2759--2769},
  publisher    = {{ACM}},
  year         = {2021},
  url          = {https://doi.org/10.1145/3459637.3482486},
  doi          = {10.1145/3459637.3482486},
  bibsource    = {dblp computer science bibliography, https://dblp.org}
}

@inproceedings{10.1145/3580305.3599331,
author = {Yang, Bei and Gu, Jie and Liu, Ke and Xu, Xiaoxiao and Xu, Renjun and Sun, Qinghui and Liu, Hong},
title = {Empowering General-purpose User Representation with Full-life Cycle Behavior Modeling},
year = {2023},
isbn = {9798400701030},
publisher = {Association for Computing Machinery},
address = {New York, NY, USA},
url = {https://doi.org/10.1145/3580305.3599331},
doi = {10.1145/3580305.3599331},
booktitle = {Proceedings of the 29th ACM SIGKDD Conference on Knowledge Discovery and Data Mining},
pages = {2908–2917},
numpages = {10},
location = {Long Beach, CA, USA},
series = {KDD '23}
}

@article{Mao_Zhu_Su_Cai_Li_Dong_2023, title={FinalMLP: An Enhanced Two-Stream MLP Model for CTR Prediction}, volume={37}, url={https://ojs.aaai.org/index.php/AAAI/article/view/25577}, DOI={10.1609/aaai.v37i4.25577}, abstractNote={Click-through rate (CTR) prediction is one of the fundamental tasks in online advertising and recommendation. Multi-layer perceptron (MLP) serves as a core component in many deep CTR prediction models, but it has been widely shown that applying a vanilla MLP network alone is ineffective in learning complex feature interactions. As such, many two-stream models (e.g., Wide&amp;Deep, DeepFM, and DCN) have recently been proposed, aiming to integrate two parallel sub-networks to learn feature interactions from two different views for enhanced CTR prediction. In addition to one MLP stream that learns feature interactions implicitly, most of the existing research focuses on designing another stream to complement the MLP stream with explicitly enhanced feature interactions. Instead, this paper presents a simple two-stream feature interaction model, namely FinalMLP, which employs only MLPs in both streams yet achieves surprisingly strong performance. In contrast to sophisticated network design in each stream, our work enhances CTR modeling through a feature selection module, which produces differentiated feature inputs to two streams, and a group-wise bilinear fusion module, which effectively captures stream-level interactions across two streams. We show that FinalMLP achieves competitive or even better performance against many existing two-stream CTR models on four open benchmark datasets and also brings significant CTR improvements during an online A/B test in our industrial news recommender system. We envision that the simple yet effective FinalMLP model could serve as a new strong baseline for future development of two-stream CTR models. Our source code will be available at MindSpore/models and FuxiCTR/model_zoo.}, number={4}, journal={Proceedings of the AAAI Conference on Artificial Intelligence}, author={Mao, Kelong and Zhu, Jieming and Su, Liangcai and Cai, Guohao and Li, Yuru and Dong, Zhenhua}, year={2023}, month={Jun.}, pages={4552-4560} }

@inproceedings{10.1145/3583780.3615134,
author = {Liu, Qidong and Yan, Fan and Zhao, Xiangyu and Du, Zhaocheng and Guo, Huifeng and Tang, Ruiming and Tian, Feng},
title = {Diffusion Augmentation for Sequential Recommendation},
year = {2023},
isbn = {9798400701245},
publisher = {Association for Computing Machinery},
address = {New York, NY, USA},
url = {https://doi.org/10.1145/3583780.3615134},
doi = {10.1145/3583780.3615134},
booktitle = {Proceedings of the 32nd ACM International Conference on Information and Knowledge Management},
pages = {1576–1586},
numpages = {11},
location = {Birmingham, United Kingdom},
series = {CIKM '23}
}

@inproceedings{10.1145/3583780.3614929,
author = {Luo, Sichun and Ma, Chen and Xiao, Yuanzhang and Song, Linqi},
title = {Improving Long-Tail Item Recommendation with Graph Augmentation},
year = {2023},
isbn = {9798400701245},
publisher = {Association for Computing Machinery},
address = {New York, NY, USA},
url = {https://doi.org/10.1145/3583780.3614929},
doi = {10.1145/3583780.3614929},
booktitle = {Proceedings of the 32nd ACM International Conference on Information and Knowledge Management},
pages = {1707–1716},
numpages = {10},
location = {Birmingham, United Kingdom},
series = {CIKM '23}
}

@inproceedings{10.1145/3292500.3330859,
author = {Lee, Hoyeop and Im, Jinbae and Jang, Seongwon and Cho, Hyunsouk and Chung, Sehee},
title = {MeLU: Meta-Learned User Preference Estimator for Cold-Start Recommendation},
year = {2019},
isbn = {9781450362016},
publisher = {Association for Computing Machinery},
address = {New York, NY, USA},
url = {https://doi.org/10.1145/3292500.3330859},
doi = {10.1145/3292500.3330859},
booktitle = {Proceedings of the 25th ACM SIGKDD International Conference on Knowledge Discovery \& Data Mining},
pages = {1073–1082},
numpages = {10},
location = {Anchorage, AK, USA},
series = {KDD '19}
}

@inproceedings{10.1145/3746252.3761574,
author = {Lu, Yusheng and Du, Zhaocheng and Li, Xiangyang and Jia, Pengyue and Wang, Yejing and Liu, Weiwen and Wang, Yichao and Guo, Huifeng and Tang, Ruiming and Dong, Zhenhua and Duan, Yongrui and Zhao, Xiangyu},
title = {Prompt Tuning as User Inherent Profile Inference Machine},
year = {2025},
isbn = {9798400720406},
publisher = {Association for Computing Machinery},
address = {New York, NY, USA},
url = {https://doi.org/10.1145/3746252.3761574},
doi = {10.1145/3746252.3761574},
booktitle = {Proceedings of the 34th ACM International Conference on Information and Knowledge Management},
pages = {5898–5906},
numpages = {9},
location = {Seoul, Republic of Korea},
series = {CIKM '25}
}

@inproceedings{wang2025personalized,
  title={Personalized Visual Content Generation in Conversational Systems},
  author={Wang, Xianquan and Du, Zhaocheng and Xu, Huibo and Yin, Shukang and Han, Yupeng and Zhu, Jieming and Zhang, Kai and Liu, Qi},
  booktitle={The Thirty-ninth Annual Conference on Neural Information Processing Systems},
  year={2025}
}

@inproceedings{10.1145/3711896.3737262,
author = {Lu, Xingyu and Wang, Jinpeng and Zhu, Jieming and Zhang, Zhicheng and Zou, Deqing and Zheng, Hai-Tao and Xia, Shu-Tao and Zhang, Rui},
title = {ROMA: Recommendation-Oriented Language Model Adaptation Using Multi-Modal Multi-Domain Item Sequences},
year = {2025},
isbn = {9798400714542},
publisher = {Association for Computing Machinery},
address = {New York, NY, USA},
url = {https://doi.org/10.1145/3711896.3737262},
doi = {10.1145/3711896.3737262},
booktitle = {Proceedings of the 31st ACM SIGKDD Conference on Knowledge Discovery and Data Mining V.2},
pages = {4670–4681},
numpages = {12},
location = {Toronto ON, Canada},
series = {KDD '25}
}

\end{document}